\documentclass{article}

\usepackage[preprint]{neurips_2026}

\usepackage[utf8]{inputenc} % allow utf-8 input
\usepackage[T1]{fontenc}    % use 8-bit T1 fonts
\usepackage{hyperref}       % hyperlinks
\usepackage{url}            % simple URL typesetting
\usepackage{booktabs}       % professional-quality tables
\usepackage{amsfonts}       % blackboard math symbols
\usepackage{nicefrac}       % compact symbols for 1/2, etc.
\usepackage{microtype}      % microtypography
\usepackage{xcolor}         % colors
\usepackage{amsmath}
\usepackage{multirow}
\usepackage{pifont}
\newcommand{\cmark}{\ding{51}}

\usepackage{graphicx}
\usepackage{pgfplots}
\pgfplotsset{compat=1.18}
\usepackage{caption}
\usepackage{wrapfig}
\usepackage{tikz}
\usepackage[table]{xcolor}
\usepackage{makecell}
\usepackage{subcaption}

\usepackage{booktabs}
\definecolor{myblue}{RGB}{232,244,255}

\DeclareRobustCommand{\geocotmark}{%
\begin{tikzpicture}[baseline=-0.55ex]
\begin{axis}[
    hide axis,
    width=0.75cm,
    height=0.25cm,
    scale only axis,
    xmin=0, xmax=1,
    ymin=0, ymax=1,
    clip=false,
]
\addplot+[
    green!60!black,
    mark=square*,
    dashed,
    thick,
    mark size=1.6pt,
    mark options={solid, fill=green!60!black, draw=green!60!black},
    mark indices={2}
] coordinates {(0,0.5) (0.5,0.5) (1,0.5)};
\end{axis}
\end{tikzpicture}%
}

\DeclareRobustCommand{\omnimark}{%
\begin{tikzpicture}[baseline=-0.55ex]
\begin{axis}[
    hide axis,
    width=0.22cm,
    height=0.25cm,
    scale only axis,
    xmin=0, xmax=1,
    ymin=0, ymax=1,
    clip=false,
]
\addplot+[
    only marks,
    mark=diamond*,
    red,
    mark size=2.2pt,
    mark options={solid, fill=red, draw=red}
] coordinates {(0.5,0.5)};
\end{axis}
\end{tikzpicture}%
}

\title{Explicit Geometric Chain-of-Thought for Vision-Language-Action in Autonomous Driving}

\author{
\textbf{
Xingtai~Gui$^{1}$
\quad
Yucheng~Zhou$^{1}$
\quad
Dongqian~Guo$^{1}$
\quad
Jiahao~Gong$^{2}$
\quad 
Feiyang~Tan$^{2}$
}
\\
\textbf{
Jianbing~Shen$^{1}$\thanks{Corresponding author.}
}
\\[0.5em]
$^{1}$SKL-IOTSC, Department of AI, University of Macau
\qquad
$^{2}$Afari Intelligent Drive
}

\begin{document}

\maketitle

\begin{abstract}
Vision-language-action~(VLA) models have emerged as a promising paradigm for autonomous driving. However, existing VLA models still suffer from a fundamental mismatch: driving actions require precise 3D geometric cues, while visual-language understanding and reasoning are largely conducted in a 2D semantic space. In this paper, we propose GeoCoTDrive, an explicit geometric chain-of-thought framework that grounds geometry in a planning-oriented manner. GeoCoTDrive follows a ``\textit{think with 2D first, drive with dedicated 3D priors}" paradigm. It first grounds 2D regions corresponding to decision-critical cues, and then retrieves localized 3D priors by sampling features from a geometric foundation model within the grounded regions. 
These localized geometric features are interleaved into the autoregressive context to support the trajectory generation. To supervise this process, we introduce planning-relevant grounding, a new region-level grounding task that focuses on local spatial cues directly affecting ego planning decisions, and construct the PlanningGrounding dataset to endow VLAs with planning-oriented grounding capability. Experiments across multiple end-to-end autonomous driving benchmarks show that GeoCoTDrive consistently improves safety-critical planning performance, demonstrating the effectiveness of the explicit geometric chain-of-thought process for VLA-based planning. Code is available at \url{https://github.com/TabGuigui/GeoCoTDrive}.

\end{abstract}

\section{Introduction}

Vision-language models~(VLMs) have been increasingly adopted in autonomous driving for question answering~\cite{yang2023llm4drive, hu2025vision, qian2024nuscenes, sima2024drivelm, marcu2024lingoqa}, scene understanding~\cite{jiang2024senna, xie2025vlms, cui2025chain, tian2025nuscenes, ding2024holistic}, and end-to-end planning~\cite{wang2025omnidrive, fu2025orion, li2025spacedrive, han2025dme, li2025recogdrive, zhou2025autovla}. These efforts collectively demonstrate that natural language can provide an effective interface for structured supervision, interpretable decision making, and planning-oriented reasoning. However, a fundamental mismatch remains in VLM-based end-to-end autonomous driving: the target action space is inherently geometric, whereas the intermediate reasoning space of VLMs is largely semantic and lacks explicit geometric cues.

This mismatch becomes especially problematic in scenarios that require fine-grained local spatial reasoning, where planning decisions are governed by localized spatial constraints. Prior works have explored several strategies to mitigate this gap. Perception-based methods\cite{wang2025omnidrive, fu2025orion, zhou2025opendrivevla} introduce 3D structural representations~\cite{sun2025sparsedrive, jiang2023vad, chen2024vadv2} or BEV-style query features~\cite{hu2023planning, liao2025diffusiondrive, gui2025trajdiff} to strengthen perception and planning relying on 3D perception supervision such as object detection or online mapping. Another line, motivated by recent advances in spatial intelligence~\cite{zheng2025learning, li2025spatial, hu2025g2vlmgeometrygroundedvision}, incorporates geometric priors into VLA models, either through explicit geometric inputs~\cite{liu2025drivepi, li2025spacedrive} or multimodal representation alignment~\cite{wang2026vggdrive, luo2026last}. However, these approaches primarily introduce geometry as global scene-level representations, rather than explicitly local and decision-critical spatial cues that directly govern planning.

\begin{figure}[t]
    \centering
    \small
    \includegraphics[width=\textwidth]{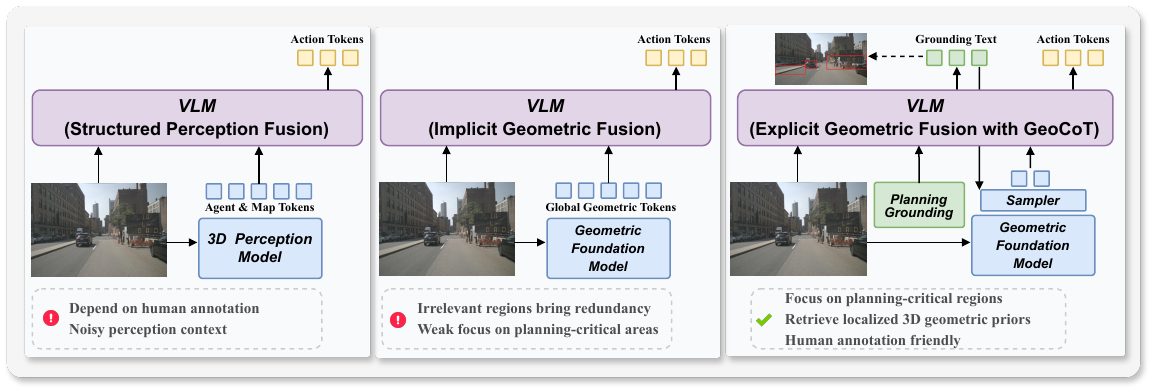}
    \caption{\textbf{Comparison of geometry integration paradigms for VLA}. (a)~Structure-perception fusion injects agent and map tokens. (b)~Geometric fusion incorporates global geometric tokens. (c)~GeoCoTDrive introduces an explicit geometric chain-of-thought pipeline: VLM first grounds planning-relevant regions, then retrieves localized 3D geometric priors, and finally interleaves the grounding text with geometry tokens to condition trajectory generation.}
    \label{fig:pipeline}
    \vspace{-4mm}
\end{figure}

Unlike prior methods that inject global geometric representations through implicit feature fusion, we propose \textit{GeoCoTDrive}, an explicit geometric chain-of-thought framework that grounds geometry in a planning-oriented manner. As shown in Fig.~\ref{fig:pipeline}, GeoCoTDrive follows a ``\textit{think with 2D first, drive with dedicated 3D priors}'' paradigm. It first leverages the reasoning capability of VLMs to identify sparse 2D regions corresponding to local decision-critical cues, then retrieves dedicated 3D geometric priors by sampling features from a geometric foundation model within the grounded regions. These geometry tokens are interleaved into the autoregressive context to support the trajectory generation.

To enable explicit planning-oriented geometry grounding, we introduce \textit{planning-relevant grounding}, a new task for end-to-end planning. Unlike conventional object-level grounding, which is typically built upon human-annotated objects with explicit semantic categories, planning-relevant grounding aims to identify region-level cues that directly affect the ego vehicle's future behavior. These cues go beyond foreground objects and include lane boundaries, conflict regions, occluded areas, and other local spatial constraints that govern planning. By shifting the grounding target from semantically defined objects to decision-critical regions, this task encourages the model to focus on the regions that are relevant to ego planning. Based on this formulation, we construct a semi-automatic annotation pipeline, yielding \textit{PlanningGrounding}, a VQA-style grounding dataset containing 146K planning-relevant grounding annotations.

Given a front-view image, a driving command, and the ego status, GeoCoTDrive first autoregressively generates planning-relevant grounding results, localizing 2D regions that capture decision-critical cues. Meanwhile, the same image is encoded by a frozen geometric foundation model to produce dense geometric features. The generated grounding coordinates are then used to sample region-level geometric features, which are projected into the language-token space and interleaved back into the autoregressive context as localized geometry tokens. Conditioned on both the 2D reasoning trace and the retrieved 3D priors, GeoCoTDrive continues autoregressive decoding to output the future ego trajectory. During training, the grounding stage is supervised by PlanningGrounding annotations, while the planning stage is supervised by expert trajectories. During inference, both stages are executed sequentially in the same autoregressive pipeline.

Extensive experiments show that GeoCoTDrive consistently improves performance across multiple end-to-end autonomous driving benchmarks. By explicitly grounding planning-relevant regions and retrieving dedicated local 3D features, GeoCoTDrive enhances both trajectory accuracy and safety-critical planning metrics.

Our main contributions are summarized as follows:

\begin{itemize}
    \item We propose GeoCoTDrive, an explicit geometric chain-of-thought framework for VLA-based end-to-end autonomous driving. It follows a ``think with 2D first, drive with dedicated 3D priors'' paradigm to bridge semantic reasoning and planning.
    
    \item We formulate planning-relevant grounding, a new region-level grounding task that focuses on local decision-critical cues directly affecting ego planning, and construct PlanningGrounding, a VQA-style grounding dataset with 146K annotations.
    
    \item We introduce an autoregressive geometry retrieval mechanism that samples localized 3D features from a geometric foundation model according to predicted grounding regions and interleaves them into the generation context for trajectory prediction.

\end{itemize}

\section{Related Work}
\subsection{VLAs for Autonomous Driving}

Recent progress in VLMs~\cite{liu2023visual, bai2023qwen, chen2024internvl, yang2025qwen3} for autonomous driving has been largely driven by language-annotated driving datasets and dedicated model designs~\cite{yang2023llm4drive, zhou2024vision, cui2025chain, hu2025vision}. Early efforts formulate driving scene understanding as visual question answering or instruction following~\cite{qian2024nuscenes, tian2025nuscenes, sima2024drivelm, marcu2024lingoqa, jiang2024senna, ding2024holistic, wang2025omnidrive, ma2024dolphins, zhou2025hermes, xu2024drivegpt4, luo2025adathinkdrive}. Building on these language-supervised tasks, recent works extend VLMs toward vision-language-action modeling. Some methods directly formulate planning as autoregressive language generation, where the model predicts high-level decisions or textual future trajectories from multimodal driving inputs~\cite{han2025dme, wang2025omnidrive, shao2024lmdrive, huang2024making, zhou2025autovla}. Other methods avoid using language models to generate precise trajectories directly, and leverage VLMs to provide high-level reasoning contexts for dedicated planning modules or action decoders~\cite{jiang2024senna, li2025recogdrive, yang2025drivemoe, fu2025orion, chen2025solve, li2026sgdrive}.

\subsection{Spatial-aware VLAs for Autonomous Driving}
Spatial awareness is increasingly recognized as a critical capability for VLMs/VLAs when downstream tasks require precise spatial decisions~\cite{chen2024spatialvlm, yang2025thinking, liu2025spatial, qu2025spatialvla}. Recent studies enhance spatial reasoning by incorporating geometric foundation models or geometry-aware representations through explicit geometric inputs~\cite{hong20233d, xu2024pointllm, sun2025geovla, li2026pointvla} or representation alignment~\cite{yuan2025depthvla, li2025spatial, hu2025g2vlmgeometrygroundedvision, zheng2025learning, li2026thinking}. These works validate the effectiveness of geometric priors in improving spatial understanding and action generation.

In autonomous driving, spatial representation is fundamental to scene understanding~\cite{ding2024holistic, sima2024drivelm}, perception~\cite{wang2022detr3d, li2024bevformer, wei2023surroundocc}, prediction~\cite{zhang2022beverse, gui2024fiptr, zheng2024occworld}, and planning~\cite{hu2023planning, chen2024vadv2, yang2025driving}. Recent VLA-based driving methods introduce human-annotated 3D supervision to learn structured scene representations~\cite{wang2025omnidrive, fu2025orion, zhou2025opendrivevla, li2026unidrivevla}. These 3D-structured representations provide important contextual cues for VLA reasoning and trajectory generation. Meanwhile, emerging methods further integrate geometry priors from pretrained foundation models~\cite{hu2024metric3d, wang2025vggt, piccinelli2025unidepthv2, lin2025depth} into VLA-based driving systems. VGGDrive~\cite{wang2026vggdrive} injects cross-view VGGT~\cite{wang2025vggt} features into VLM hidden states through a hierarchical adaptive enabler. SpaceDrive~\cite{li2025spacedrive} converts UniDepth-derived 3D coordinates~\cite{piccinelli2025unidepthv2} into universal positional encodings for spatially aware visual and coordinate tokens. LaST-VLA~\cite{luo2026last} distills geometric priors from VGGT into continuous latent CoT states for physically grounded planning. Despite their effectiveness, these methods primarily introduce global geometry priors without 2D reasoning derivation. In contrast, GeoCoTDrive explicitly constructs a planning-oriented reasoning path from identifying 2D decision-critical regions to retrieving localized 3D geometry.

\section{Method}

\subsection{Preliminaries}
\paragraph{Vision-Language-Action Model.}~In vision-language-action for autonomous driving, the core objective is to predict future ego trajectories from multimodal driving contexts based on pretrained VLMs. Given visual observations $I$ and language contexts $X$ including driving command, ego status, and textual prompts, the model predicts a future ego trajectory $\tau=\{(x_t,y_t,h_t)\}_{t=1}^{T}$ where $t\in\{1,\dots,T\}$ denotes the prediction step within the future horizon, and $\tau_t=(x_t,y_t,h_t)$ denotes the trajectory state at step $t$. A driving VLA models the planning distribution as:
\begin{equation}
P_{\theta}(\tau \mid X, I)=\prod_{t=1}^{T} P_{\theta}(\tau_t \mid \tau_{<t}, X, I),
\end{equation}

\paragraph{Geometric Foundation Model.}~Geometric foundation models provide dense spatial representations learned from large-scale geometry-aware pretraining, offering complementary geometric priors. We instantiate the geometric feature extractor $f_{\mathrm{geo}}$ with two foundation models: Depth Anything V3~\cite{lin2025depth} and VGGT~\cite{wang2025vggt}. DepthAnythingV3 provides dense monocular geometry cues with strong robustness across diverse visual conditions, while VGGT offers 3D-aware latent representations learned from large-scale geometric pretraining. We use the intermediate backbone features from these frozen geometry models as geometric priors for geometric chain-of-thought. Specifically, we sample region-level geometric tokens from the feature map $G=f_{\mathrm{geo}}(I)$.

\begin{figure*}[t]
    \centering
    \includegraphics[width=\textwidth]{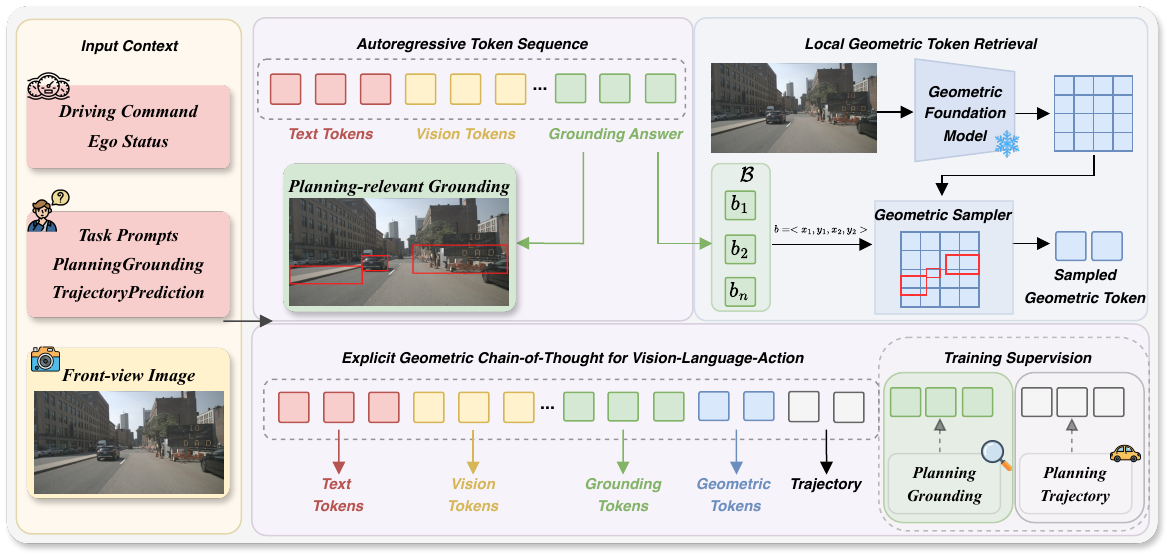}
    \caption{\textbf{Overview of GeoCoTDrive.} The model first grounds planning-critical 2D regions, retrieves localized geometry tokens from a geometric foundation model, and then interleaves them into the autoregressive context for trajectory generation.}
    \label{fig:main}
\end{figure*}

\subsection{GeoCoTDrive}
GeoCoTDrive models VLA as an explicit geometric chain-of-thought process. Given the textual prompt $X$ and the front-view image $I \in \mathbb{R}^{H \times W \times 3}$, the model first leverages the reasoning ability inherited from pretrained VLMs to generate a set of planning-relevant regions:
\begin{equation}
    \mathcal{B}=\{b_i\}_{i=1}^{N} \sim P_{\theta}(\mathcal{B}\mid X,I),
\end{equation}
where each region $b_i$ corresponds to a local decision-critical cue. The generated regions provide an explicit intermediate reasoning trace that specifies where geometry should be retrieved. GeoCoTDrive then samples a set of localized geometric tokens from the dense geometric feature map:
\begin{equation}
    \mathcal{Z}_{\mathcal{B}} = \mathrm{Sample}(G,\mathcal{B}),
\end{equation}
where $G$ denotes the geometric feature map extracted from the same image. After the grounding output, the geometric tokens are interleaved into the autoregressive context, enabling GeoCoTDrive to continue decoding the trajectory conditioned on both 2D grounding and retrieved 3D priors:
\begin{equation}
    P_{\theta}(\tau \mid X,I,\mathcal{B},\mathcal{Z}_{\mathcal{B}})
    =
    \prod_{t=1}^{T}
    P_{\theta}(\tau_t \mid \tau_{<t}, X,I,\mathcal{B},\mathcal{Z}_{\mathcal{B}}).
\end{equation}
Concretely, the grounding output is a set of 2D bounding boxes, where each box is represented by normalized coordinates:
\begin{equation}
    b_i=(x_i^{\min}, y_i^{\min}, x_i^{\max}, y_i^{\max}), \qquad i=1,\ldots,N .
\end{equation}

Given the dense geometric feature map $G \in \mathbb{R}^{H_g \times W_g \times C_g}$ extracted by the geometric foundation model, GeoCoTDrive samples geometric features from each grounding box. A regular grid with $K$ sampling points is constructed inside each box, and bilinear interpolation is applied over $G$:
\begin{equation}
    \tilde{\mathbf{Z}}_i
    =
    S(G, b_i)
    \in \mathbb{R}^{K \times C_g},
\end{equation}
where $S(\cdot)$ denotes the region-based sampler. The resulting $\tilde{\mathbf{Z}}_i$ contains $K$ localized geometric features from the $i$-th planning-relevant region.

The sampled features are then projected into the LLM space through a lightweight alignment head:
\begin{equation}
    \mathbf{Z}_i
    =
    f_{\mathrm{align}}(\tilde{\mathbf{Z}}_i)
    \in \mathbb{R}^{K \times d},
\end{equation}
where $f_{\mathrm{align}}$ is implemented as a MLP projection head and $d$ is the hidden dimension of the pretrained VLM. The geometry tokens from all grounded regions are concatenated as:
\begin{equation}
    \mathcal{Z}_{\mathcal{B}}
    =
    [\mathbf{Z}_1;\mathbf{Z}_2;\cdots;\mathbf{Z}_N]
    \in \mathbb{R}^{(N \times K) \times d}.
\end{equation}
The concatenated geometry tokens $\mathcal{Z}_{\mathcal{B}}$ are interleaved into the autoregressive context after the grounding output. This enables the model to continue generation with explicit access to localized 3D evidence corresponding to the planning-critical regions it has just identified.

\subsection{Training Objective}

GeoCoTDrive is optimized with a standard autoregressive language modeling objective under teacher forcing. For each training sample, we construct a multimodal sequence by placing the grounding response, inserted geometry tokens, and trajectory response within the same autoregressive context. Let $\mathbf{s}=\{s_l\}_{l=1}^{L}$ denote the resulting sequence, the training objective is defined as:
\begin{equation}
    \mathcal{L}_{\mathrm{AR}}
    =
    -\frac{1}{|\mathcal{M}|}
    \sum_{l \in \mathcal{M}}
    \log P_{\theta}(s_l \mid s_{<l}),
\end{equation}
where $\mathcal{M}$ denotes the set of supervised token positions, including both grounding and trajectory responses. Tokens corresponding to user prompts, image placeholders, and inserted geometry features are masked in the loss function.

\section{Planning-oriented Grounding Dataset}

\begin{figure*}[t]
    \centering
    \includegraphics[width=\textwidth]{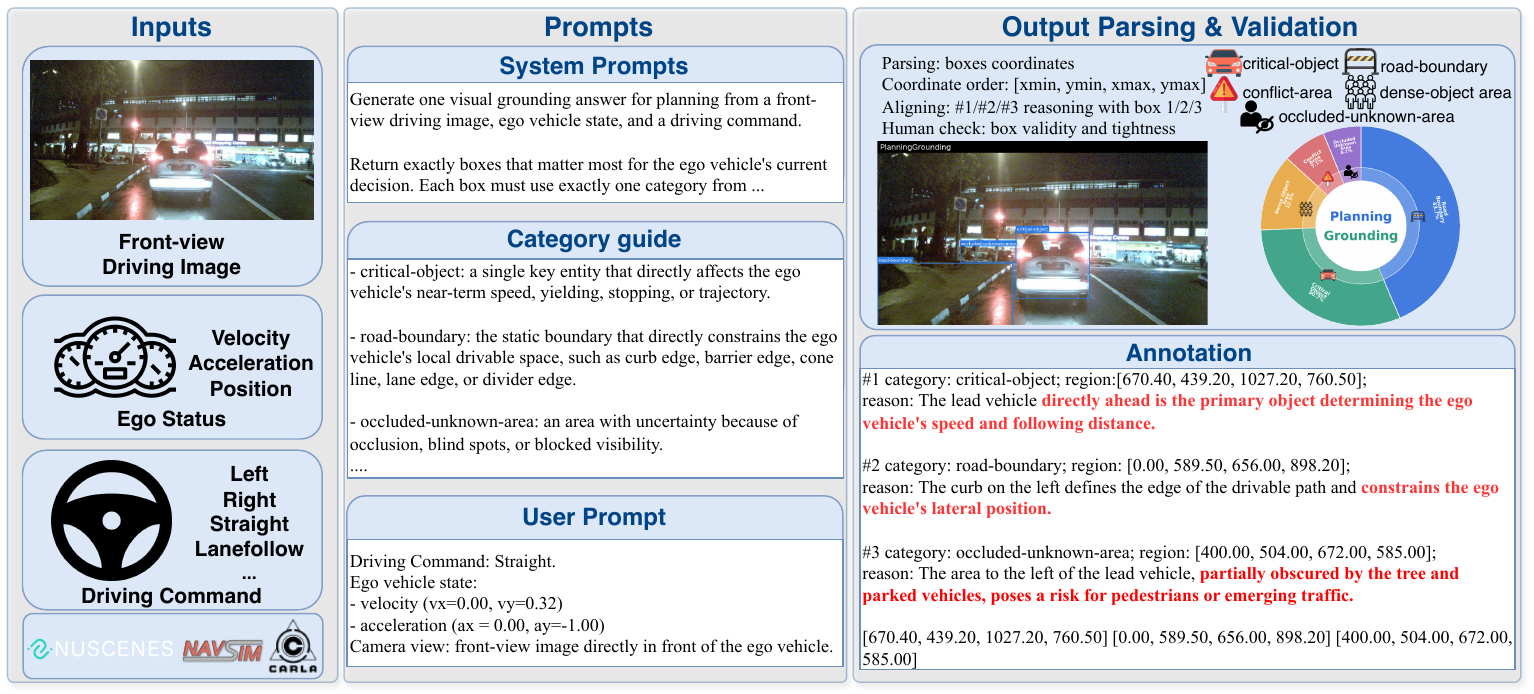}
    \caption{\textbf{Overview of the PlanningGrounding data construction pipeline.} The pipeline generates planning-oriented region annotations from driving scenes by identifying decision-critical visual cues. An annotation example and the overall category distribution of the dataset are provided.}
    \label{fig:flowchart}

\end{figure*}

End-to-end driving requires models to identify not only semantic objects, but also local regions that directly affect ego planning. Conventional object-level grounding is category-driven and may overlook planning-critical spatial cues such as conflict areas, occluded regions, and dense agent clusters. To address this limitation, we formulate planning-relevant grounding, which aims to localize a compact set of decision-critical regions conditioned on the driving scene.

As shown in Fig.~\ref{fig:flowchart}, each annotation sample is constructed from a front-view driving image, ego status, and a high-level driving command. A planning-oriented prompt is used to guide Gemini-3.1~\cite{team2023gemini} to identify the regions most relevant to the ego decision and assign each region to a planning-related category. The generated outputs are parsed into bounding boxes and category-reasoning pairs, followed by automatic format validation and human verification to ensure box validity and planning relevance. To cover diverse driving scenarios and benchmark settings, the raw driving data is collected from nuScenes~\cite{caesar2020nuscenes}, NAVSIM~\cite{dauner2024navsim}, and Bench2Drive~\cite{jia2024bench2drive}.

Following this semi-automatic annotation pipeline, we construct PlanningGrounding, a VQA-style grounding dataset with 146K annotations. The annotations cover five planning-related categories: \textit{critical-object}, \textit{road-boundary}, \textit{conflict-area}, \textit{occluded-unknown-area}, and \textit{dense-object-area}. The first two categories are closely related to object- and map-centric annotations commonly used in existing driving VLM/VLA datasets~\cite{wang2025omnidrive, fu2025orion}, and therefore constitute a large proportion of the dataset. Beyond these conventional cues, PlanningGrounding further provides region-level annotations for planning-critical structures, including conflict-area, dense-object-area, and occluded-unknown-area. In addition, 0.6K manually refined samples are provided as the test set of the PlanningGrounding benchmark. For more information about this dataset, please refer to the Appendix.

% -------------------- Experiment

\begin{table*}[t]
\centering
\footnotesize
\caption{\textbf{Planning results on nuScenes valset.}
Best and second-best results among VLM/VLA-based methods are highlighted. The result follows the evaluation protocol of OmniDrive~\cite{wang2025omnidrive}.}
\setlength{\tabcolsep}{3.8pt}
\begin{tabular}{lcc|cccc|cccc|cccc}
\toprule
\multirow{2}{*}{Method} & \multicolumn{2}{c|}{Ego Status} 
& \multicolumn{4}{c|}{L2 (m) $\downarrow$}
& \multicolumn{4}{c|}{Collision (\%) $\downarrow$}
& \multicolumn{4}{c}{Intersection (\%) $\downarrow$} \\
\cmidrule(lr){2-3} \cmidrule(lr){4-7} \cmidrule(lr){8-11} \cmidrule(l){12-15}
& BEV & Planner
& 1s & 2s & 3s & Avg.
& 1s & 2s & 3s & Avg.
& 1s & 2s & 3s & Avg. \\
\midrule

\multicolumn{15}{l}{\textit{Traditional / modular paradigm}} \\

ST-P3\cite{hu2022st}                  & --     & --     & 1.59 & 2.64 & 3.73 & 2.65 & 0.69 & 3.62 & 8.39 & 4.23 & 2.53 & 8.17 & 14.40 & 8.37 \\
UniAD\cite{hu2023planning}                  & \cmark & \cmark & 0.20 & 0.42 & 0.75 & 0.46 & 0.02 & 0.25 & 0.84 & 0.37 & 0.20 & 1.33 & 3.24  & 1.59 \\
VAD-Base\cite{jiang2023vad}               & \cmark & \cmark & 0.17 & 0.34 & 0.60 & 0.37 & 0.04 & 0.27 & 0.67 & 0.33 & 0.21 & 2.13 & 5.06  & 2.47 \\
AD-MLP\cite{li2024ego}                 & --     & \cmark & 0.15 & 0.32 & 0.59 & 0.35 & 0.00 & 0.27 & 0.85 & 0.37 & 0.27 & 2.52 & 6.60  & 2.93 \\
BEV-Planner++\cite{li2024ego}          & \cmark & \cmark & 0.16 & 0.32 & 0.57 & 0.35 & 0.00 & 0.29 & 0.73 & 0.34 & 0.35 & 2.62 & 6.51  & 3.16 \\
UAD\cite{guo2025end}                    & \cmark & \cmark & 0.13 & 0.28 & 0.48 & 0.30 & 0.00 & 0.12 & 0.55 & 0.22 & 0.10 & 0.80 & 2.38  & 1.13 \\

\midrule
\multicolumn{15}{l}{\textit{VLM / VLA-based paradigm}} \\

RDA-Driver\cite{huang2024making}
& \cmark & \cmark
& 0.23 & 0.73 & 1.54 & 0.80
& \textbf{0.00} & 0.13 & 0.83 & 0.32
& -- & -- & -- & -- \\

OmniDrive\cite{wang2025omnidrive}
& \cmark & \cmark
& \textbf{0.14} & \underline{0.29} & 0.55 & 0.33
& \textbf{0.00} & 0.13 & 0.78 & 0.30
& 0.56 & 2.48 & 5.96 & 3.00 \\

ORION\cite{fu2025orion}
& \cmark & --
& 0.17 & 0.31 & 0.55 & 0.34
& 0.05 & 0.25 & 0.80 & 0.37
& -- & -- & -- & -- \\

VGGDrive\cite{wang2026vggdrive}
& -- & \cmark
& \textbf{0.14} & \textbf{0.28} & \textbf{0.51} & \textbf{0.31}
& \underline{0.02} & \underline{0.10} & 0.55 & 0.22
& 0.63 & 2.27 & \underline{4.02} & 2.31 \\

LaST-VLA\cite{luo2026last}
& -- & \cmark
& 0.17 & 0.33 & 0.64 & 0.38
& \textbf{0.00} & 0.11 & \underline{0.42} & \underline{0.18}
& -- & -- & -- & -- \\

SpaceDrive\cite{li2025spacedrive}
& -- & \cmark
& \underline{0.15} & \underline{0.29} & \textbf{0.51} & \underline{0.32}
& 0.04 & 0.18 & 0.49 & 0.23
& \textbf{0.22} & \textbf{0.80} & \textbf{2.79} & \textbf{1.27} \\

\rowcolor{myblue}
\textbf{GeoCoTDrive}
& -- & \cmark
& \textbf{0.14} & \underline{0.29} & \underline{0.53} & \underline{0.32}
& \textbf{0.00} & \textbf{0.07} & \textbf{0.27} & \textbf{0.11}
& \underline{0.37} & \underline{1.44} & 4.25 & \underline{2.02} \\

\bottomrule
\end{tabular}
\vspace{-4mm}
\label{tab:nuscenes}
\end{table*}

\section{Experiments}
\subsection{Experimental setup}

\paragraph{Evaluation metrics.}~We evaluate the end-to-end planning performance on three autonomous driving benchmarks: nuScenes~\cite{caesar2020nuscenes}, NAVSIM~\cite{dauner2024navsim} and Bench2Drive~\cite{jia2024bench2drive}. For nuScenes, we report open-loop planning performance using L2 distance error, collision rate, and intersection rate. 
For NAVSIM, we adopt the Predictive Driver Model Score (PDMS) as the closed-loop planning metric, which consists of five sub-scores: no-at-fault collisions (NC), drivable area compliance (DAC), time-to-collision (TTC), comfort (Comf.), and ego progress (EP). 
For Bench2Drive, we follow the official closed-loop evaluation protocol and report driving score (DS), success rate (SR). 
Together, these metrics provide a comprehensive assessment of planning accuracy, safety, comfort, and progress.

For planning-relevant grounding, we further evaluate GeoCoTDrive on the manually refined PlanningGrounding test set, which contains 0.6K samples and measures the model's ability to localize decision-critical regions before trajectory generation. The grounding performance is evaluated using Recall@N, average precision~(AP), and mean intersection-over-union~(mIoU).

\paragraph{Training details.}~GeoCoTDrive is instantiated with different base VLMs across benchmarks to follow the standard configurations of prior baseline methods. For nuScenes and Bench2Drive, we adopt LLaVA-1.5-7B as the base VLM, following OmniDrive~\cite{wang2025omnidrive} and ORION~\cite{fu2025orion}. For NAVSIM, we use Qwen2.5-VL as the base VLM to align with recent VLA-based planning methods~\cite{li2025recogdrive, luo2026last}. For the trajectory decoder, nuScenes and NAVSIM use autoregressive text decoding, where future trajectory coordinates are generated as language action tokens. On Bench2Drive, we follow the diffusion planner design in ORION~\cite{fu2025orion} to produce executable trajectories. This heterogeneous setup demonstrates that GeoCoTDrive can be integrated with different VLM backbones and trajectory-generation heads, highlighting its generality across diverse autonomous driving benchmarks.

For geometric feature extraction, we evaluate DepthAnythingV3~\cite{lin2025depth} and VGGT~\cite{wang2025vggt}, which provide dense depth-aware and 3D-aware feature maps for localized geometry retrieval. The geometric foundation model is kept frozen. GeoCoTDrive is trained in two stages. In Stage I, we update the vision encoder and LLM with full QA training to adapt the VLM to autonomous-driving scenarios. In Stage II, we freeze the vision encoder and optimize the LLM together with the geometric alignment head, enabling localized geometric features to be aligned with the language space and interleaved into the planning context. More training details are provided in the Appendix.

\vspace{-3mm}

\begin{table*}[t]
\centering
\footnotesize
\caption{\textbf{Planning results on NAVSIM navtest split}. The result follows the evaluation protocol of official NAVSIMv1\cite{dauner2024navsim} with non-reactive simulation.}

\label{tab:navsim-main}
\begin{tabular}{lcc|cccccc}
\toprule
\multirow{2}{*}{Method} & \multirow{2}{*}{Base Model} & \multirow{2}{*}{Sensor}
& \multicolumn{6}{c}{Closed-loop Metrics ($\uparrow$)} \\
\cmidrule(l){4-9}
& & & NC & DAC & EP & TTC & Comf. & PDMS \\
\midrule
\multicolumn{9}{l}{\textit{Traditional / modular paradigm}} \\
TransFuser~\cite{chitta2022transfuser} & - & Image+LiDAR & 97.8 & 92.6 & 78.9 & 92.0 & 99.9 & 83.8 \\
PARA-Drive~\cite{weng2024drive} & - & Image & 97.9 & 92.4 & 79.3 & 93.0 & 99.8 & 84.0 \\
Hydra-MDP~\cite{li2024hydra} & - & Image+LiDAR & 98.3 & 96.0 & 78.7 & 94.6 & 100.0 & 86.5 \\
DiffusionDrive~\cite{liao2025diffusiondrive} & - & Image+LiDAR & 98.2 & 96.2 & 82.2 & 94.7 & 100.0 & 88.1 \\
\midrule
\multicolumn{9}{l}{\textit{VLM / VLA-based paradigm}} \\
Baseline (Qwen) & Qwen2.5-VL-3B & Image & 98.0 & 94.0 & 80.4 & 94.4 & 99.9 & 85.9 \\
AutoVLA~\cite{zhou2025autovla} & Qwen2.5-VL-3B & Image & 96.9 & 92.4 & 75.8 & 88.1 & 99.9 & 80.5 \\
ReCogDrive~\cite{li2025recogdrive} & InternVL3-8B & Image & 98.3 & 95.1 & \underline{81.1} & 94.3 & 100.0 & 86.8 \\
LaST-VLA~\cite{luo2026last} &  InternVL3-8B & Image & \textbf{98.7} & \underline{95.4} & 80.5 & \textbf{95.6} & 100.0 & \underline{87.3} \\
\rowcolor{myblue}
\textbf{GeoCoTDrive} & Qwen2.5-VL-3B & Image & \underline{98.4} & \textbf{95.6} & \textbf{81.6} & \underline{95.0} & 100.0 & \textbf{87.6} \\
\bottomrule
\end{tabular}
\vspace{-2mm}
\end{table*}

\subsection{Main results}

\paragraph{Planning.}~Table~\ref{tab:nuscenes} reports the open-loop planning results on nuScenes. GeoCoTDrive achieves competitive performance among VLA-based methods and shows favorable safety-related performance. Table~\ref{tab:navsim-main} further evaluates GeoCoTDrive on the NAVSIM navtest set. Compared with the Qwen2.5-VL baseline, GeoCoTDrive achieves a modest PDMS improvement and improves safety-related sub-metrics. Among recent VLA-based planners, GeoCoTDrive remains competitive despite using a smaller VLM backbone. Table~\ref{tab:bench2drive} shows that GeoCoTDrive improves over both the perception-centric and the global geometry fusion method with the Bench2Drive closed-loop evaluation.

\begin{table*}[t]
\centering
\begin{minipage}[t]{0.42\textwidth}
\centering
\footnotesize
\caption{\textbf{Closed-loop Planning results on Bench2Drive.}~Driving Score (DS) and Success Rate (SR) are reported.}
\label{tab:bench2drive}
\vspace{1mm}
\begin{tabular}{l|cc}
\toprule
\multirow{2}{*}{Method} & \multicolumn{2}{c}{Closed-loop Metrics ($\uparrow$)} \\
\cmidrule(l){2-3}
& DS & SR \\
\midrule
ORION~\cite{fu2025orion} & 77.74 & 54.62 \\
SpaceDrive~\cite{li2025spacedrive} & 78.02 & 55.11 \\

\rowcolor{myblue}
\textbf{GeoCoTDrive} & \textbf{78.68} & \textbf{55.42} \\
\bottomrule
\end{tabular}

\end{minipage}
\hfill
\begin{minipage}[t]{0.54\textwidth}
\centering
\renewcommand{\arraystretch}{1.1}
\footnotesize
\setlength{\tabcolsep}{3pt}
\caption{\textbf{Grounding results on the PlanningGrounding test set.} We report Recall at different IoU thresholds, AP, and mIoU.}
\label{tab:plangrounding}
\vspace{1mm}
\begin{tabular}{lccccc}
\toprule
Method & R@0.3$\uparrow$ & R@0.5$\uparrow$ & R@0.7$\uparrow$ & AP$\uparrow$ & mIoU$\uparrow$ \\
\midrule
LLaVA1.5-7B & 41.39 & 24.49 & 9.25 & 25.04 & 27.98 \\
Qwen2.5-VL-3B & 46.27 & 33.86 & 9.84 & 30.61 & 30.46 \\
OmniDrive & 52.56 & 41.03 & 10.90 & 34.83 & 32.30 \\
\rowcolor{myblue}
\textbf{GeoCoTDrive} & \textbf{56.82} & \textbf{42.73} & \textbf{21.04} & \textbf{39.89} & \textbf{38.93} \\
\bottomrule
\end{tabular}
\end{minipage}
\vspace{-4mm}

\end{table*}

\paragraph{Grounding.}~We evaluate grounding performance on the PlanningGrounding test set shown in Table~\ref{tab:plangrounding}. Driving-specialized models outperform general VLM baselines, suggesting that autonomous-driving instruction tuning helps localize planning-relevant cues. GeoCoTDrive achieves the best results across all metrics, with particularly clear gains under stricter IoU thresholds. These improvements indicate that GeoCoTDrive benefits from the PlanningGrounding dataset to localize decision-critical regions more precisely, providing more reliable spatial anchors for localized geometry retrieval and trajectory generation.

\begin{table*}[t]
    \centering
    \scriptsize
    \renewcommand{\arraystretch}{1.08}
    \caption{\textbf{Ablation study on geometry integration strategies.} We compare the vanilla baseline, global geometry fusion, and the proposed GeoCoT.}
    \label{tab:ablation}
    \resizebox{\textwidth}{!}{
    \begin{tabular}{lccccc|ccc}
        \toprule
        \multirow{2}{*}{Method} 
        & \multicolumn{5}{c|}{NAVSIM} 
        & \multicolumn{3}{c}{nuScenes} \\
        \cmidrule(lr){2-6} \cmidrule(lr){7-9}
        & NC  
        & DAC  
        & TTC  
        & EP 
        & PDMS 
        & L2 
        & CR 
        & Intersection \\
        \midrule
        Baseline 
        & 98.0 & 94.0 & 94.4 & 80.4 & 85.9
        & 0.16 / 0.32 / 0.58 
        & 0.04 / 0.11 / 0.52 
        & 0.64 / 2.46 / 5.31 \\
        Global 
        & 97.9 & 94.1 & 94.2 & 80.2 & 85.8
        & 0.16 / 0.31 / 0.57 
        & 0.02 / 0.17 / 0.56 
        & 0.48 / 2.34 / 4.90 \\
        GeoCoT 
        & \textbf{98.4} & \textbf{95.6} & \textbf{95.0} & \textbf{81.6} & \textbf{87.6}
        & \textbf{0.14} / \textbf{0.29} / \textbf{0.53} 
        & \textbf{0.00} / \textbf{0.07} / \textbf{0.27} 
        & \textbf{0.37} / \textbf{1.44} / \textbf{4.25} \\
        \bottomrule
    \end{tabular}
    \vspace{-8mm}
    }
\end{table*}

\begin{figure}[t]
    \centering
    \small
    \vspace{-2mm}
    \includegraphics[width=\textwidth]{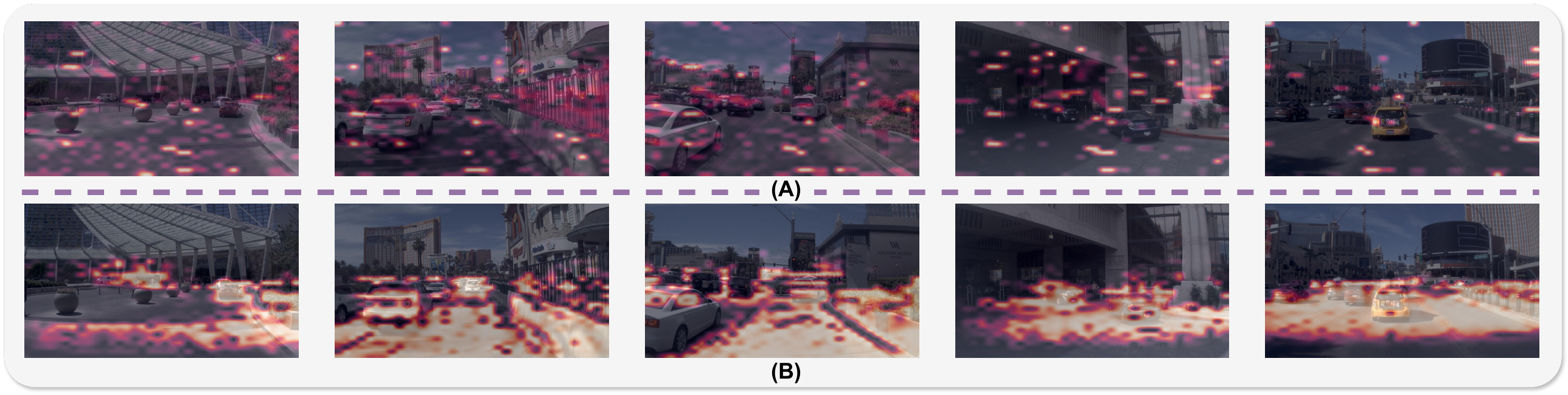}
    \caption{\textbf{Visualization of geometry token response with different fusion strategies.}~(A) Global geometry tokens interleaved (B) GeoCoT retrieved geometric tokens interleaved.}
    \label{fig:compare}
    \vspace{-4mm}
\end{figure}

\subsection{Ablation study}

\begin{table}[t]
    \centering
    \footnotesize
    \renewcommand{\arraystretch}{1.08}
    \caption{\textbf{Comparison of different grounding annotation sources for GeoCoTDrive.} ``Perception label'' uses the human-annotated perception boxes.}
    \label{tab:label_source_ablation}
    \setlength{\tabcolsep}{8pt}
    \resizebox{0.9\linewidth}{!}{
    \begin{tabular}{lccc}
        \toprule
        Method & L2 & CR & Intersection \\
        \midrule
        GeoCoTDrive w/ Perception Label & 0.17 / 0.30 / 0.55 & 0.02 / 0.12 / 0.39 & 0.50 / 2.20 / 4.98 \\
        GeoCoTDrive w/ PlanningGrounding    & \textbf{0.14} / \textbf{0.29} / \textbf{0.53} & \textbf{0.00} / \textbf{0.07} / \textbf{0.27} & \textbf{0.37} / \textbf{1.44} / \textbf{4.25} \\
        \bottomrule
    \end{tabular}}
    \vspace{-3mm}
\end{table}

\paragraph{Ablation on geometry integration.}
Table~\ref{tab:ablation} compares different strategies for incorporating geometric information. Directly interleaving global geometry tokens yields limited gains, suggesting that indiscriminate geometry injection may introduce redundant or weakly relevant spatial cues. In contrast, GeoCoTDrive achieves more consistent improvements by retrieving localized geometric priors from planning-relevant grounded regions. These results indicate that geometry is more effective when conditioned on decision-critical regions, rather than injected as a global auxiliary stream.

We provide a qualitative comparison of geometric responses under different geometry insertion strategies. We calculated the attention output of the action token and geometric token. As shown in Fig.~\ref{fig:compare}~(A), directly inserting global geometry tokens fails to focus on regions that are truly relevant to the driving decision. In contrast, Fig.~\ref{fig:compare} shows that GeoCoTDrive retrieves geometry tokens from planning-grounded regions and yields responses that are more aligned with decision-critical cues. This visualization further supports the advantage of planning-oriented geometry retrieval over indiscriminate global geometry injection.

\paragraph{Effect of grounding annotation source.}~Table~\ref{tab:label_source_ablation} compares different grounding annotation sources for GeoCoTDrive. Perception labels provide comprehensive human-annotated scene elements, but they are primarily designed for general perception and are not necessarily aligned with planning. In contrast, PlanningGrounding focuses on regions that are directly relevant to driving decisions.
The consistent improvements, especially on safety-related metrics, indicate that GeoCoTDrive benefits from localized geometry retrieval guided by planning-relevant grounding.

\begin{table}[t]
\centering
\footnotesize
\caption{\textbf{Effect of geometric foundation model.} Models are loaded from official checkpoints.}
\label{tab:geometric_ablation}
\setlength{\tabcolsep}{11pt}
\resizebox{0.85\linewidth}{!}{
\begin{tabular}{l c | c c c c c}
\toprule
Method & Geometric Model & NC & DAC & TTC & EP & PDMS \\
\midrule
Baseline & -- & 98.0 & 94.0 & 94.4 & 80.4 & 85.9 \\
GeoCoTDrive & VGGT & 98.3 & 95.3 & 94.6 & \textbf{81.7} & 87.2 \\
GeoCoTDrive & DA3-LARGE & \textbf{98.4} & \textbf{95.6} & \textbf{95.0} & 81.6 & \textbf{87.6} \\
\bottomrule
\end{tabular}}
\vspace{-2mm}
\end{table}

\begin{wrapfigure}{r}{0.58\textwidth}
    \vspace{-3mm}
    \centering
    \begin{minipage}[t]{0.48\linewidth}
    \centering
    \begin{tikzpicture}
    \begin{axis}[
        width=\linewidth,
        height=3.6cm,
        xlabel={mIoU},
        ylabel={Collision Rate Avg.},
        xmin=0.25, xmax=0.40,
        ymin=0.10, ymax=0.32,
        ytick={0.1,0.2,0.3},
        grid=major,
        tick label style={font=\scriptsize},
        label style={font=\scriptsize},
    ]
    \addplot[
        green!60!black,
        mark=square*,
        dashed,
        thick,
        mark size=2.0pt,
        mark options={solid, fill=green!60!black, draw=green!60!black}
    ] coordinates {(0.2558,0.14) (0.3271,0.12) (0.3853,0.11)};

    \addplot[
        only marks,
        mark=diamond*,
        red,
        mark size=2.2pt,
        mark options={solid, fill=red, draw=red}
    ] coordinates {(0.323,0.30)};
    \end{axis}
    \end{tikzpicture}
    \end{minipage}
    \hfill
    \begin{minipage}[t]{0.48\linewidth}
    \centering
    \begin{tikzpicture}
    \begin{axis}[
        width=\linewidth,
        height=3.6cm,
        xlabel={mIoU},
        ylabel={Intersection Avg.},
        xmin=0.25, xmax=0.40,
        ymin=1.9, ymax=3.2,
        grid=major,
        tick label style={font=\scriptsize},
        label style={font=\scriptsize},
    ]
    \addplot[
        green!60!black,
        mark=square*,
        dashed,
        thick,
        mark size=2.0pt,
        mark options={solid, fill=green!60!black, draw=green!60!black}
    ] coordinates {(0.2558,2.30) (0.3271,2.10) (0.3853,2.02)};

    \addplot[
        only marks,
        mark=diamond*,
        red,
        mark size=2.2pt,
        mark options={solid, fill=red, draw=red}
    ] coordinates {(0.3230,3.00)};
    \end{axis}
    \end{tikzpicture}
    \end{minipage}

    \vspace{-1mm}
    \caption{\textbf{Effect of grounding quality on downstream driving behavior.}
    Lower values indicate better safety performance. 
    \protect\geocotmark~denotes GeoCoTDrive with perturbed grounding quality, and
    \protect\omnimark~denotes OmniDrive.}
    \label{fig:grounding_planning_correlation}
    \vspace{-4mm}
\end{wrapfigure}
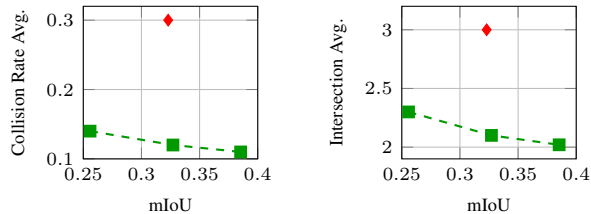

\paragraph{Effect of grounding quality on downstream planning.}~
We perturb the generated grounding results during GeoCoT to analyze their impact on planning. As shown in Fig.~\ref{fig:grounding_planning_correlation}, higher grounding mIoU consistently improves safety-related metrics, indicating that accurate planning-relevant grounding retrieves more reliable localized geometry. Although OmniDrive achieves comparable grounding mIoU to an intermediate GeoCoTDrive setting, its downstream safety metrics remain worse, suggesting that the gains come from explicitly coupling 2D grounding evidence with localized 3D priors.
\begin{figure}[!ht]
    \centering
    \small
    \includegraphics[width=1.0\textwidth]{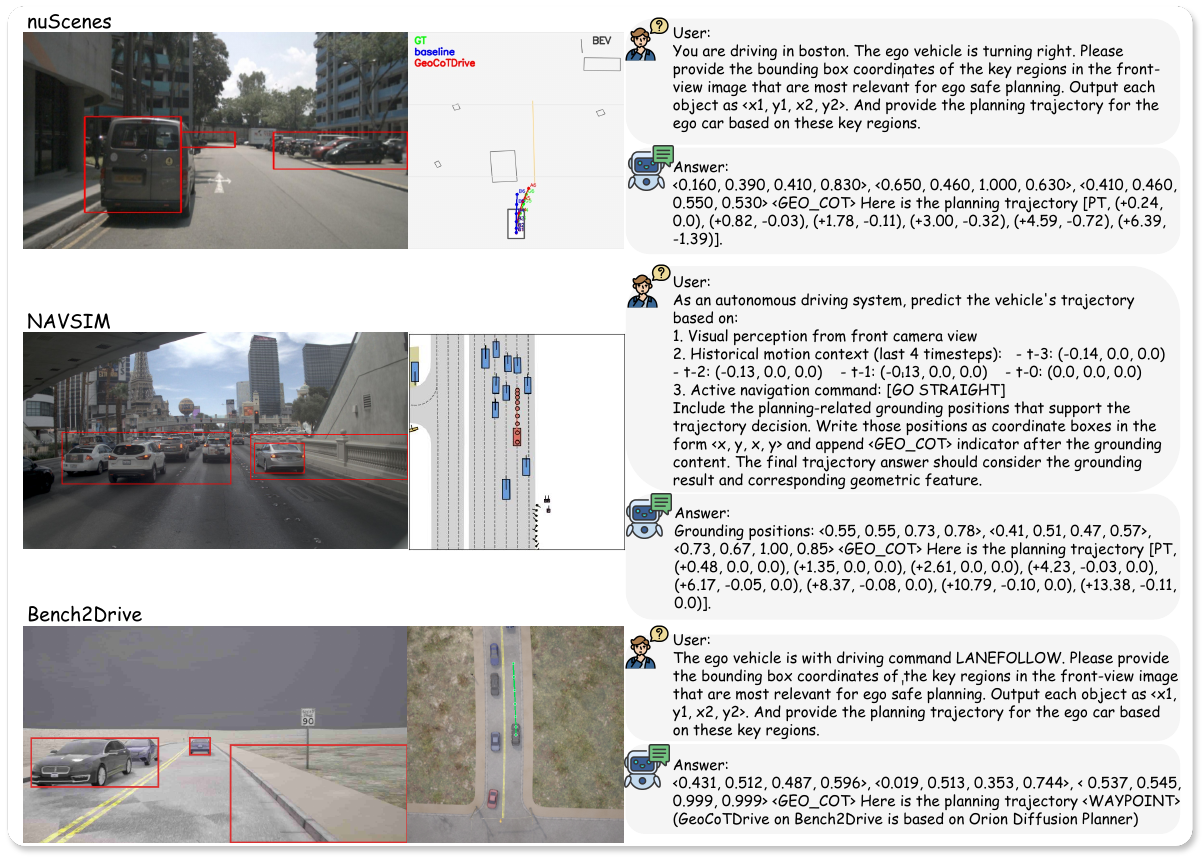}
    \caption{\textbf{Qualitative planning results of GeoCoTDrive on nuScenes, NAVSIM and Bench2Drive.}}
    \label{fig:planning}

\end{figure}

\paragraph{Effect of geometric foundation model.}
Table~\ref{tab:geometric_ablation} studies the impact of different geometric foundation models. Both VGGT and DA3-LARGE improve over the baseline, suggesting that localized geometry is generally beneficial for planning. DA3-LARGE achieves slightly better overall performance, while VGGT obtains a marginally higher EP. These results indicate that GeoCoTDrive is not tied to a specific geometry backbone and can benefit from diverse pretrained geometric priors.
\vspace{-1mm}

\subsection{Qualitative Results}

Fig.~\ref{fig:planning} visualizes the qualitative planning results of GeoCoTDrive on nuScenes, NAVSIM, and Bench2Drive. Compared with the baseline, GeoCoTDrive shows stronger geometry-aware safety behavior. The visualization results across nuScenes, NAVSIM, and Bench2Drive further show that the proposed GeoCoT pipeline is consistently effective under evaluation protocols and driving scenarios. Additional qualitative results are provided in the Appendix.

\section{Conclusion}

We present GeoCoTDrive, an explicit geometric chain-of-thought framework for VLA-based end-to-end autonomous driving. GeoCoTDrive first grounds planning-relevant 2D regions, retrieves localized 3D geometric priors from a frozen geometric foundation model, and interleaves these geometry tokens into the autoregressive context for trajectory generation. We further introduce PlanningGrounding, a VQA-style grounding dataset that provides region-level supervision for decision-critical spatial cues. Experiments on nuScenes, NAVSIM, and Bench2Drive demonstrate that GeoCoTDrive improves planning performance, highlighting the importance of coupling planning-oriented 2D reasoning with localized 3D geometric priors for safer autonomous driving.

% \newpage

\bibliographystyle{plain}
\bibliography{ref}

\newpage

\clearpage
% ------------------------------------------ supp ----------------
\appendix
\vskip8pt
\noindent\textbf{\textit{\LARGE Appendix}}
\vskip8pt

The Appendix is organized as follows:
\begin{itemize}
    \item Sec.~\ref{sec:app_vqa_grounding}: comparison with existing driving VQA datasets on grounding supervision.
    \item Sec.~\ref{sec:app_plangrounding}: detailed analysis of the proposed PlanningGrounding dataset.
    \item Sec.~\ref{sec:app_implementation_geocot}: detailed training and inference implementation of GeoCoTDrive.
    \item Sec.~\ref{sec:app_more_ablation}: more ablation studies.
    \item Sec.~\ref{sup:qualitative}: more qualitative results.
    \item Sec.~\ref{sup:limitations}: limitations of GeoCoTDrive.
\end{itemize}

\section{Autonomous Driving VQA}
\label{sec:app_vqa_grounding}

\begin{table*}[h]
\centering
\small
\caption{\textbf{Comparison with existing driving VQA datasets on grounding supervision.}}
\label{tab:vqa_grounding_compare}
\begin{tabular}{p{0.18\textwidth} p{0.415\textwidth} p{0.31\textwidth}}
\toprule
Dataset & Main supervision & Grounding granularity \\
\midrule
NuScenes-QA~\cite{qian2024nuscenes}
& Template-based VQA from structured scene graphs.
& Object-level attributes and relations derived from 3D detection annotations. \\
\midrule
DriveLM~\cite{sima2024drivelm}
& Graph Visual Question Answering for perception, prediction, planning, behavior, and motion.
& Object-level graph nodes and interaction-level reasoning. \\
\midrule
LingoQA~\cite{marcu2024lingoqa}
& Video-based driving QA with natural-language answers and explanations.
& Mainly language-level supervision without explicit box-level grounding. \\
\midrule
OmniDrive~\cite{wang2025omnidrive}
& Counterfactual QA using simulated trajectories, expert trajectories, 3D objects, and map elements.
& 3D object/map-centric detection and trajectory-level consequence reasoning. \\
\midrule

PlanningGrounding
& Command-conditioned grounding of planning-relevant visual regions.
& Region-level 2D boxes covering objects, boundaries, conflict zones, occlusions, and dense object groups. \\
\bottomrule
\end{tabular}
\end{table*}

Recent autonomous-driving VQA datasets have significantly advanced language-based scene understanding, reasoning, and planning. However, the grounding supervision of these datasets is usually designed around object-level perception annotations, rather than explicit region-level cues related to the ego vehicle's planning demand. We summarize the main differences in  Table~\ref{tab:vqa_grounding_compare}.

Unlike conventional object-centric grounding, PlanningGrounding is designed to identify the regions that most directly affect the ego vehicle's planning decision. This distinction is important because the planning decisions are often constrained not only by individual traffic participants, but also by local road boundaries, conflict driving zones, or uncertain regions. Therefore, PlanningGrounding complements existing driving VQA resources by providing explicit, command-aware grounding supervision for planning-oriented reasoning.

\section{Analysis of PlanningGrounding}
\label{sec:app_plangrounding}

PlanningGrounding is constructed to supervise explicit planning-oriented visual grounding. Given a front-view driving image, driving command, and optional ego status, the task is to localize the most decision-critical regions that affect the planning decision. Each grounding region is assigned one of five categories: \emph{critical-object}, \emph{road-boundary}, \emph{conflict-area}, \emph{occluded-unknown-area}, and \emph{dense-object-area}. These categories cover both object-level and region-level spatial cues required for safe trajectory generation.

\begin{itemize}
    \item \textbf{Critical object} denotes a single key entity that directly affects the ego vehicle's near-term speed, yielding behavior, stopping decision, or trajectory selection. Typical examples include a leading vehicle requiring deceleration, a pedestrian near the ego path, a cyclist around the turning area, a traffic cone blocking the lane, or a traffic light/sign that immediately constrains the maneuver. An object is annotated as critical only when its influence is directly tied to the current command and local planning decision.
    \item \textbf{Road boundary} refers to an immediate static boundary that constrains the ego vehicle's local drivable space. This category includes lane edges, curbs, barriers, dividers, cone lines, road edges, and other visible structures that define the lateral feasible region.
    \item \textbf{Conflict area} denotes a compact local region where the ego vehicle's intended path may interact with other traffic flows or potential agent motions, such as a merge point, crossing zone, or turn conflict entrance. Unlike critical object, this category describes a spatial zone rather than a single entity, encouraging the model to reason about where potential interactions may occur.
    \item \textbf{Occluded unknown area} denotes a region with meaningful uncertainty caused by occlusion, blind spots, or blocked visibility. Such regions commonly appear behind parked vehicles, large trucks, roadside buildings, vegetation, or sharp turns. Although no explicit object may be visible, these regions can still require conservative planning, such as slowing down before entering a blind spot or preparing to yield.
    \item \textbf{Dense object area} represents a compact cluster of objects whose collective presence affects planning. Examples include a group of pedestrians, a queue of vehicles, a cluster of cones, or multiple parked vehicles narrowing the drivable corridor. This category is used when the planning-relevant cue is better represented by a region containing multiple nearby entities rather than by a single object.
\end{itemize}
 
The above categories are not intended to provide exhaustive perception labels. Instead, they define a sparse set of command-relevant regions that directly influence near-term planning. Unlike generic object detection or visual grounding, PlanningGrounding selects annotation targets according to planning relevance, covering not only salient objects but also road boundaries, conflict zones, occluded areas, and dense object regions that constrain safe trajectory generation.

\begin{table*}[!ht]
\centering
\small
\setlength{\tabcolsep}{5pt}
\caption{\textbf{Dataset composition of PlanningGrounding.}
The table summarizes the number of grounding QA samples from each source dataset and the distribution of five planning-oriented grounding categories. ``K'' denotes thousands of samples.}
\label{tab:plangrounding_composition}
\begin{tabular}{l c c c c c c}
\toprule
\multirow{2}{*}{Source Dataset} 
& \multirow{2}{*}{\# Samples} 
& \multicolumn{5}{c}{Category Distribution (\%)} \\
\cmidrule(lr){3-7}
&
& \makecell{Critical\\Object}
& \makecell{Road\\Boundary}
& \makecell{Conflict}
& \makecell{Occluded \\ Unknown}
& \makecell{Dense \\ Object} \\
\midrule
nuScenes 
& 28K 
& 30.68 
& 43.65
& 7.05
& 6.14 
& 12.45 \\

NAVSIM 
& 102K 
& 34.47
& 34.05 
& 6.86
& 3.99
& 20.60 \\

Bench2Drive 
& 16K 
& 38.31 
& 47.16 
& 2.99 
& 3.48 
& 8.05 \\
\midrule
Total 
& 146K 
& 34.16 
& 37.33 
& 6.47 
& 4.35 
& 17.66 \\
\bottomrule
\end{tabular}
\end{table*}

Table~\ref{tab:plangrounding_composition} summarizes the source composition and category distribution of PlanningGrounding. Considering the diversity of driving scenarios, data sources, and evaluation protocols, PlanningGrounding is constructed from three widely studied autonomous-driving planning benchmarks: nuScenes, NAVSIM, and Bench2Drive. This combination enables PlanningGrounding to cover a broad range of planning-relevant cues. 

\begin{figure}[!t]
    \centering
    \small
    \includegraphics[width=\textwidth]{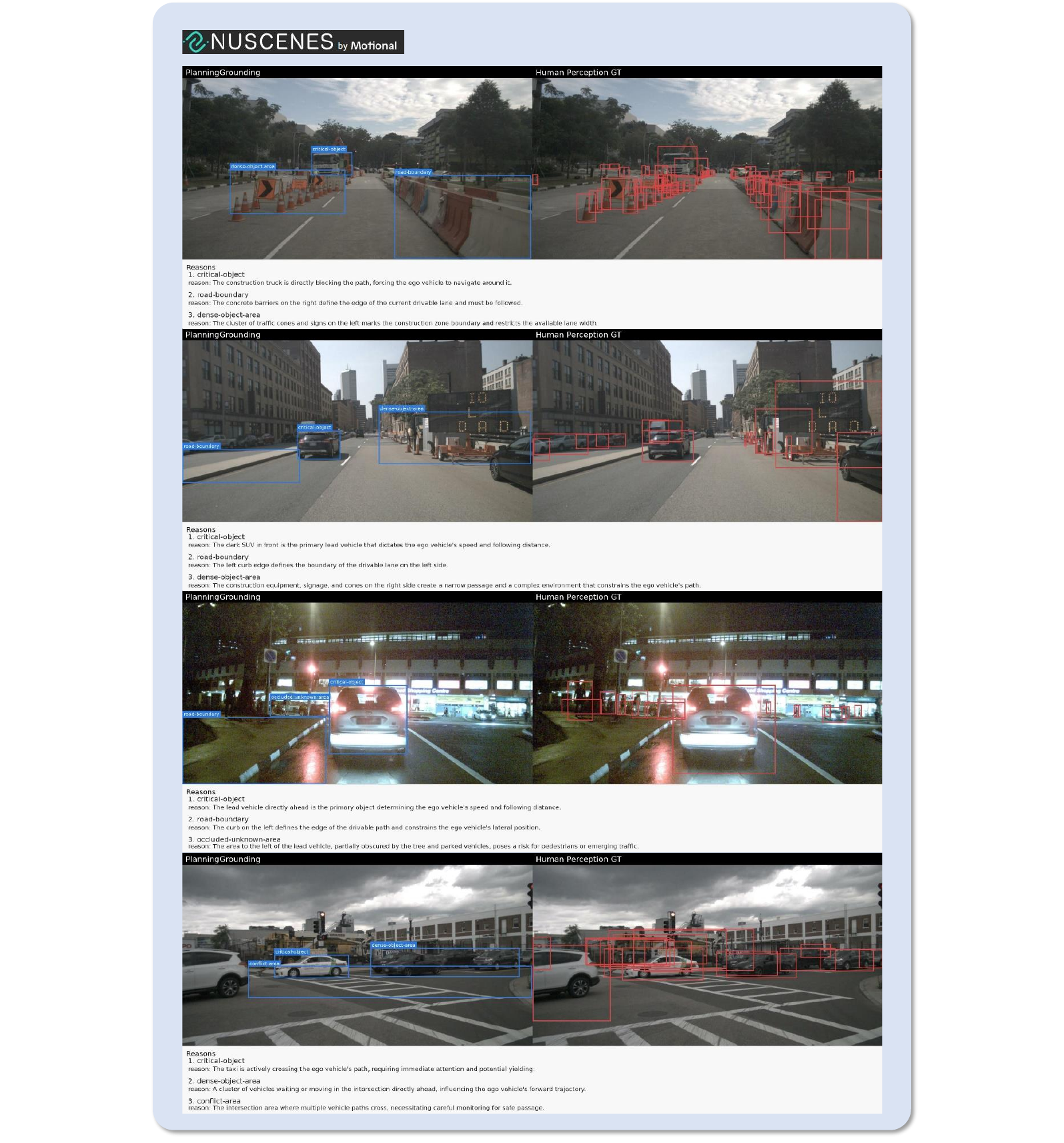}
    \vspace{-5pt}
    \caption{\textbf{PlanningGrounding-nuScenes dataset.}~Human annotations are shown for comparison.}
    \label{fig:plangrounding_examples_nuscenes}
    \vspace{-3mm}
\end{figure}

\begin{figure}[!ht]
    \centering
    \small
    \includegraphics[
        width=0.78\textwidth,
        trim=0 0 0 3.5cm,
        clip
    ]{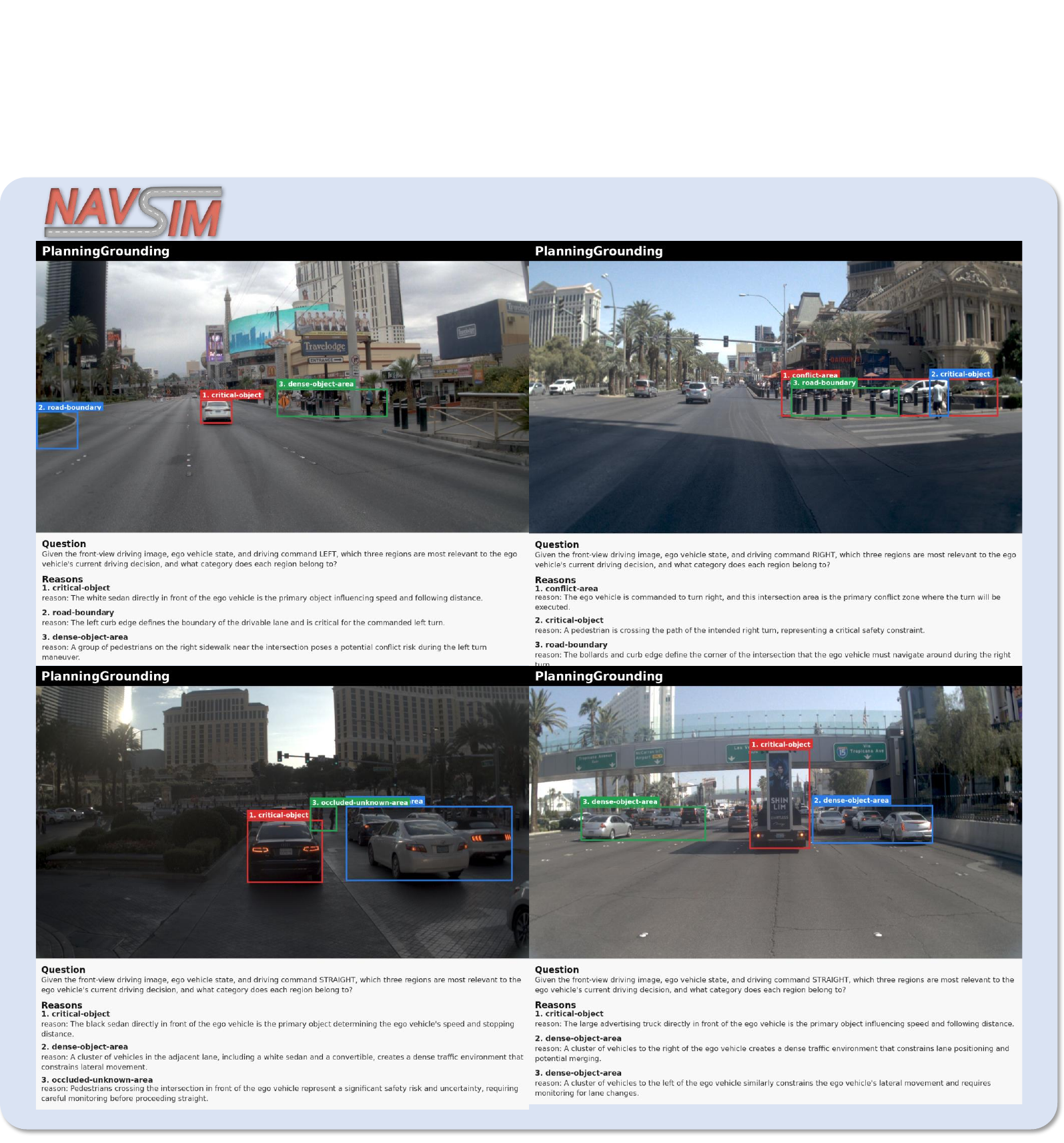}
    \includegraphics[
        width=0.78\textwidth,
        trim=0 0 0 3cm,
        clip
    ]{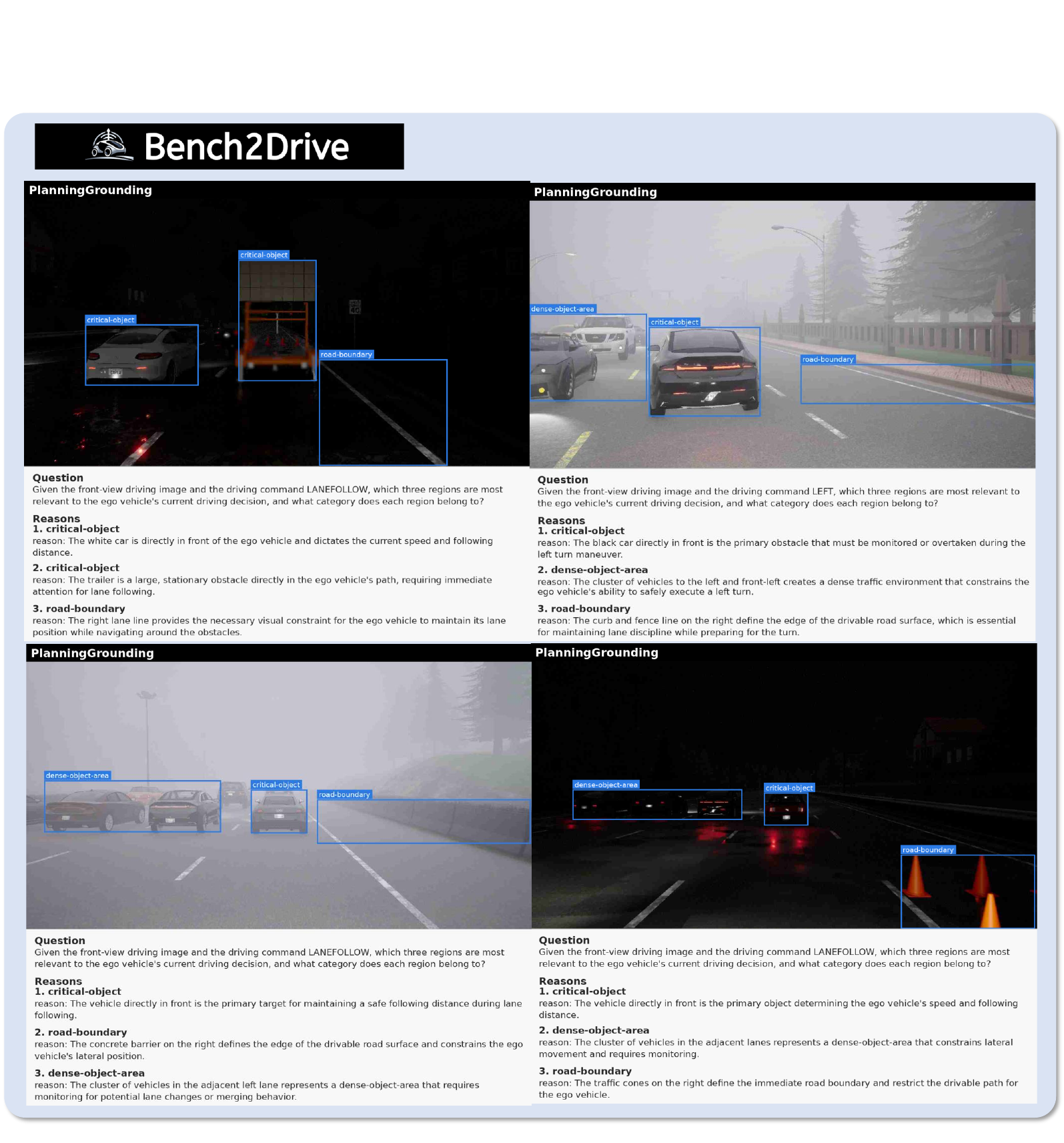}

    \caption{\textbf{PlanningGrounding-NAVSIM} and \textbf{PlanningGrounding-Bench2Drive} dataset examples.}
    \label{fig:plangrounding_examples}
    \vspace{-4mm}
\end{figure}

In Fig.~\ref{fig:plangrounding_examples_nuscenes}, we show examples from PlanningGrounding-nuScenes and compare the original human-annotated perception labels with our planning-relevant grounding annotations. Fig.~\ref{fig:plangrounding_examples} further provides examples from PlanningGrounding-NAVSIM and PlanningGrounding-Bench2Drive.

\section{Implementation Details}
\label{sec:app_implementation_geocot}

\paragraph{VLM backbone.}
GeoCoTDrive adopts different VLM backbones according to the benchmark setting. For nuScenes and Bench2Drive, LLaVA-1.5-7B is used as the base VLM, while Qwen2.5-VL-3B is used for NAVSIM. Across all settings, the model takes the front-view image, ego status, and high-level navigation command as input, and autoregressively generates planning-relevant grounding results followed by the planning answer.

\paragraph{Geometric foundation model.}
Two geometric foundation models are used to extract geometric priors: Depth Anything V3 with the METRIC-LARGE configuration and the vanilla version of VGGT. Both models are kept frozen. For each model, the feature map from the last backbone layer is used as the source of geometric tokens. Given a grounding region, localized geometric features are sampled from this feature map and projected into the language embedding space.

\paragraph{Two-stage training.}
\begin{table}[t]
\centering
\footnotesize
\setlength{\tabcolsep}{4pt}
\caption{\textbf{Two-stage training structure of GeoCoTDrive.}
Stage-I adapts the vision-language backbone to autonomous-driving QA and planning knowledge, while Stage-II trains GeoCoT planning.}
\label{tab:training_structure}
\begin{tabular}{l c c c c}
\toprule
Stage 
& Trainable Modules 
& Objective 
& Epochs 
& Peak LR \\
\midrule
Stage-I 
& Vision + LLM 
& Full QA training 
& 3 
& $4\times10^{-5}$ \\

Stage-II 
& LLM + Geometric Aligner
& GeoCoT 
& 3 
& $2\times10^{-5}$ \\
\bottomrule
\end{tabular}
\vspace{-3mm}
\end{table}

The training process contains two stages. The first stage performs full QA training to adapt the VLM to autonomous-driving knowledge. Specifically, nuScenes uses OmniDrive together with PlanningGrounding, NAVSIM uses NAVSIM-Traj together with PlanningGrounding, and Bench2Drive uses B2D-Chat together with PlanningGrounding. In this stage, both the vision encoder and the LLM are updated, allowing the model to acquire driving-related visual understanding and planning knowledge. The model is trained for 3 epochs with a cosine learning-rate scheduler, where the peak learning rate is set to $4\times10^{-5}$.

The second stage performs geometry-interleaved planning training. Planning answers and grounding annotations are merged into a single-turn conversation, where the model is required to first generate planning-relevant grounding results and then predict the planning trajectory. A special token \texttt{<GEO\_COT>} is inserted after the grounding output as a placeholder for geometric reasoning. During training, the ground-truth grounding boxes are used as the sampler input to retrieve localized geometric features. The sampled geometric tokens are then interleaved into the autoregressive sequence at the \texttt{<GEO\_COT>} position, enabling subsequent trajectory prediction to condition on localized 3D priors. This stage is also trained for 3 epochs with a cosine learning-rate scheduler, where the peak learning rate is set to $2\times10^{-5}$. For the Bench2Drive benchmark, we follow the planner implementation of ORION. We remove the detection and mapping head, and use the VLM tokens and localized geometric tokens from the front-view image as conditions for the planner. All experiments are conducted with 8 A800 GPUs.

\paragraph{Inference.}
The inference process follows the same question format as training. The model first autoregressively generates grounding results. Once the \texttt{<GEO\_COT>} token is generated, the predicted grounding boxes are parsed and used to sample localized geometric features from the geometric foundation model. These resulting geometric tokens are interleaved into the context, after which the model continues autoregressive decoding to generate the final trajectory prediction.

\begin{table}[t]
\centering
\footnotesize
\setlength{\tabcolsep}{4pt}
\caption{\textbf{Ablation on the sampling grid size for localized geometry retrieval.}
The grid size denotes the number of sampled geometric features within each grounded region.}
\label{tab:grid_size_ablation}
\begin{tabular}{c c c c c c}
\toprule
Grid Size 
& Geo. Tokens 
& NC 
& DAC
& TTC 
& PDMS \\
\midrule
$1 \times 1$ & 1  & 98.2 & 94.6 & 94.5 & 86.4 \\
$2 \times 2$ & 4  & 98.3 & 95.1 & 94.6 & 87.0 \\
$4 \times 4$ & 16 & 98.5 & 95.4 & 95.0 & 87.3 \\
$5 \times 5$ & 25 & 98.5 & 95.4 & 95.0 & 87.3 \\
\bottomrule
\end{tabular}
\vspace{-3mm}
\end{table}

\begin{table}[t]
\centering
\footnotesize
\caption{\textbf{Ablation on training stages.}
Stage-I denotes full QA training, and Stage-II denotes geometry-interleaved planning training.}
\label{tab:training_stage_ablation}
\begin{tabular}{cccccc}
\toprule
Stage-I QA & Stage-II GeoCoT & NC & DAC & TTC & PDMS \\
\midrule
$\checkmark$ & $\times$ & 98.0 & 94.0 & 94.4 & 85.9 \\
$\times$ & $\checkmark$ & 98.3 & 94.7 & 94.5 & 86.6 \\
$\checkmark$ & $\checkmark$ & 98.5 & 95.4 & 95.0 & 87.3 \\
\bottomrule
\end{tabular}
\vspace{-3mm}
\end{table}

\begin{table}[t]
    \centering
    \footnotesize
    \caption{\textbf{Performance on turning scenarios in NAVSIM.} Values in parentheses indicate absolute gains over the baseline.}
    \label{tab:navsim_turning}
    \setlength{\tabcolsep}{5pt}
    \begin{tabular}{llcccc}
        \toprule
        Scenario & Method & PDMS & NC
        & DAC & EP \\
        \midrule
        Left turn
        & Baseline
        & 83.96 & 98.34 & 91.04 & 78.63 \\
        & GeoCoTDrive
        & 86.66 (+2.70)
        & 99.04 (+0.70)
        & 93.80 (+2.76)
        & 80.78 (+2.15) \\
        \midrule
        Right turn
        & Baseline
        & 80.10 & 97.30 & 89.84 & 73.23 \\
        & GeoCoTDrive
        & 82.32 (+2.22)
        & 97.61 (+0.31)
        & 91.80 (+1.96)
        & 74.35 (+1.12) \\
        \bottomrule
    \end{tabular}
    \vspace{-3mm}
\end{table}

\begin{figure}[!ht]
    \centering
    \small
    \includegraphics[width=1.0\textwidth]{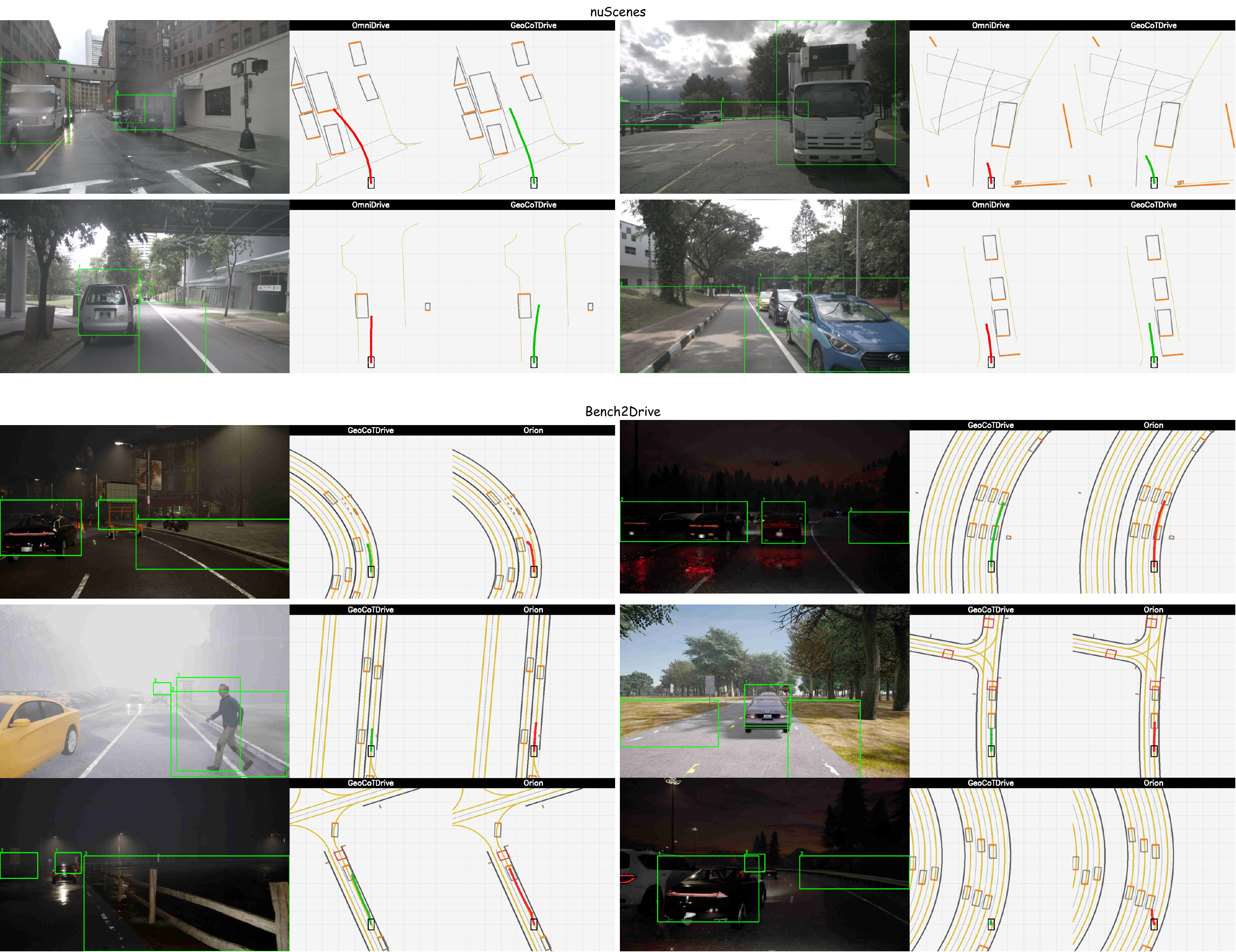}
    \caption{\textbf{Qualitative planning results on nuScenes and Bench2Drive.} GeoCoTDrive is compared with the corresponding baseline methods.}
    \label{fig:planning_supp}

\end{figure}

\section{Additional Ablation Studies}
\label{sec:app_more_ablation}

\paragraph{Effect of sampling grid size.}~Table~\ref{tab:grid_size_ablation} studies the sampling grid size for localized geometry retrieval. A small grid may miss local structures, while an overly dense grid introduces redundancy. The results show that a moderate grid size achieves the best trade-off between geometric detail and token efficiency.

\paragraph{Effect of training stages.}~Table~\ref{tab:training_stage_ablation} studies the contribution of the two-stage training strategy. Stage-I alone provides basic autonomous-driving QA and planning knowledge, while Stage-II alone enables geometry-interleaved planning. Combining both training stages achieves the best planning performance across all metrics, indicating that driving knowledge adaptation and localized geometry conditioning are complementary for end-to-end planning.

\paragraph{Scenario-wise Analysis}~We evaluate GeoCoTDrive on NAVSIM turning scenarios, where planning requires reasoning about road geometry and surrounding agents. As shown in Table~\ref{tab:navsim_turning}, GeoCoTDrive improves PDMS for both left and right turns, with gains across NC, DAC, and EP. These results support the effectiveness of grounded geometric priors in planning under local spatial constraints.

\section{Additional Qualitative Results}
\label{sup:qualitative}

Fig.~\ref{fig:planning_supp} presents additional qualitative planning comparison. Each example pairs a front-view image with BEV visualizations of the road layout, surrounding agents, and predicted trajectories. The examples span curved roads, dynamic intersections, and low-visibility conditions. GeoCoTDrive first grounds regions that influence driving decisions, such as nearby vehicles, pedestrians, and drivable areas. It then incorporates geometric priors from these regions into planning to generate safety-aware trajectories that account for road boundaries and potential conflicts with surrounding agents.

\section{Limitations}
\label{sup:limitations}

GeoCoTDrive primarily relies on front-view images without explicit multi-view or temporal modeling, limiting its ability to account for side and rear traffic, prolonged occlusions, and interaction histories. Localized geometry retrieval also depends on grounding accuracy. The incorrect or incomplete regions may yield less relevant geometric cues and impair trajectory prediction. Future work will extend GeoCoTDrive to multi-view and temporal settings and improve grounding robustness.

% \clearpage
% \input{checklist.tex}

\end{document}